\documentclass[10pt, twocolumn, letterpaper]{article}

\usepackage{iccv}              

\newcommand{\datasetname}{ Subjects200K }
\newcommand{\pname}{ OminiControl }

\usepackage{color}
\usepackage{colortbl}
\usepackage{multirow}
\usepackage{makecell}
\usepackage{wrapfig}
\usepackage{listings}
\usepackage{tabulary}
\definecolor{iccvblue}{rgb}{0.21,0.49,0.74}
\definecolor{cvprblue}{rgb}{0.21,0.49,0.74}
\usepackage[pagebackref, breaklinks, colorlinks, allcolors=iccvblue]{hyperref}

\def\paperID{1247} 
\def\confName{ICCV}
\def\confYear{2025}

\title{OminiControl: Minimal and Universal Control for Diffusion Transformer}

\author{ Zhenxiong Tan \quad Songhua Liu \quad Xingyi Yang \quad Qiaochu Xue \quad
Xinchao Wang \\ National University of Singapore\\
{ \tt \small \{zhenxiong, songhua.liu, xyang, e1352520\}@u.nus.edu xinchao@nus.edu.sg }
}

\begin{document}
    \twocolumn[{ \maketitle \begin{center}\centering \vspace{-0.5cm} \captionsetup{type=figure} \includegraphics[width=0.9\textwidth]{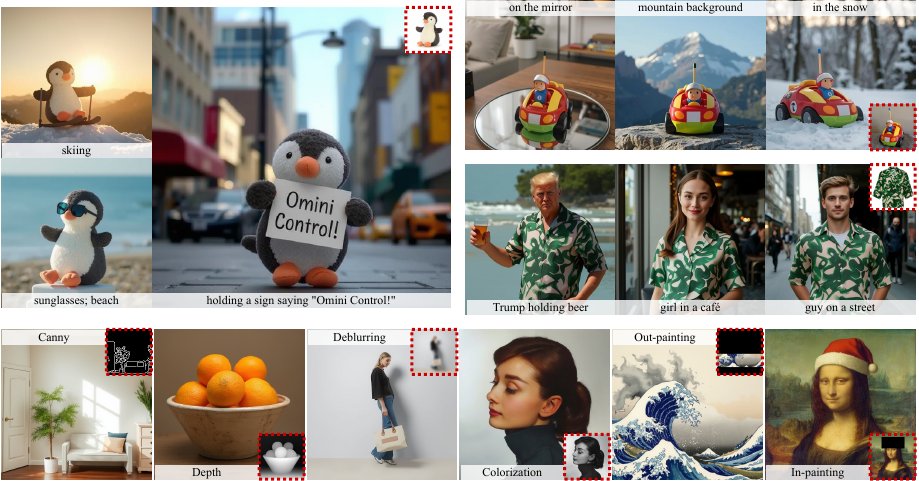} \vspace{0cm} \caption{ Results of our \pname\ on both subject-driven generation (top) and spatially-aligned tasks (bottom). The small images in red boxes show the input conditions. }\end{center} }]

    \begin{abstract}

    We present OminiControl, a novel approach that rethinks how image
    conditions are integrated into Diffusion Transformer (DiT) architectures.
    Current image conditioning methods either introduce substantial parameter
    overhead or handle only specific control tasks effectively, limiting their practical
    versatility. OminiControl addresses these limitations through three key
    innovations: (1) a minimal architectural design that leverages the DiT's own
    VAE encoder and transformer blocks, requiring just 0.1\% additional parameters;
    (2) a unified sequence processing strategy that combines condition tokens
    with image tokens for flexible token interactions; and (3) a dynamic position
    encoding mechanism that adapts to both spatially-aligned and non-aligned control
    tasks. Our extensive experiments show that this streamlined approach not
    only matches but surpasses the performance of specialized methods across
    multiple conditioning tasks. To overcome data limitations in subject-driven generation,
    we also introduce Subjects200K, a large-scale dataset of identity-consistent
    image pairs synthesized using DiT models themselves. This work demonstrates
    that effective image control can be achieved without architectural
    complexity, opening new possibilities for efficient and versatile image generation
    systems.
\end{abstract}

    \section{Introduction}
\label{sec:intro}

The ability to generate high-quality images with precise user control remains a central
challenge in computer vision. While diffusion models~\cite{ho2020denoising, rombach2021highresolution}
have significantly advanced image generation, surpassing traditional GAN-based approaches~\cite{goodfellow2020generative}
in both quality and diversity, fine-grained control remains problematic. Text-conditioned
models~\cite{rombach2021highresolution, podell2023sdxl, esser2024scaling, blackforestlabs_flux}
serve as the primary controllable generation paradigm, yet fundamentally lack
the capacity to specify exact spatial details and visual attributes that users
often need. To address this limitation, some works have explored image-based
control methods~\cite{zhang2023adding, ye2023ip, mou2024t2i, zhao2024uni,
li2024blip, shi2024instantbooth, xiao2024fastcomposer, li2024photomaker,
qin2023unicontrol, wei2023elite, li2025controlarcontrollableimagegeneration, sun2024anycontrolcreateartworkversatile,
he2024dynamiccontroladaptiveconditionselection, bar2023multidiffusion,lin2024ctrlxcontrollingstructureappearance,
hua2023dreamtunersingleimagesubjectdriven, he2025anystoryunifiedsinglemultiple,
li2023blip, rout2024semantic, xiao2024fastcomposer, wang2024instantid}, enabling
users to specify their intentions through reference images or visual hints,
offering more precise guidance than text alone~\cite{rombach2021highresolution}.

However, current image control methods face several key challenges: First, existing
methods require substantial architectural modifications with dedicated control modules~\cite{zhang2023adding,
ye2023ip, pan2023kosmos}. Second, these approaches show clear task bias -- they
typically work well for either spatially aligned controls (e.g., edge-guided or depth-guided
generation~\cite{zhang2023adding,mou2024t2i,zhao2024uni,qin2023unicontrol, li2025controlarcontrollableimagegeneration,
sun2024anycontrolcreateartworkversatile, he2024dynamiccontroladaptiveconditionselection,
bar2023multidiffusion,lin2024ctrlxcontrollingstructureappearance}) or spatially
unaligned ones (e.g., style transfer or subject-driven generation~\cite{ye2023ip,
li2024photomaker, li2024blip, shi2024instantbooth,
hua2023dreamtunersingleimagesubjectdriven}\cite{he2025anystoryunifiedsinglemultiple,
ma2024characteradapterpromptguidedregioncontrol}), but rarely both. Third,
current approaches are built primarily for UNet architectures~\cite{ronneberger2015u}.
These approaches yield suboptimal results when applied to newer Diffusion Transformer
(DiT) models~\cite{peebles2023scalable} due to fundamental architectural differences,
despite the latter's superior generation capabilities~\cite{blackforestlabs_flux,
chen2023pixart, esser2024scaling}.

These challenges, particularly within the emerging paradigm of Diffusion Transformer
models, motivate us to rethink the fundamental approach to image control. We propose
\textbf{\textit{\pname}}, an \textbf{omni}-capable yet \textbf{minimal} framework
that achieves effective and flexible \textbf{control}.

For minimal architecture, {\pname} employs a parameter reuse strategy that leverages
DiT's existing components—particularly its VAE encoder and transformer blocks—which
already possess the necessary capabilities to process visual control signals. By
reusing these components with minimal fine-tuning, our approach dramatically reduces
parameter overhead while maintaining control effectiveness, requiring only 0.1\%
additional parameters compared to the base model.

To achieve omni-capability across diverse tasks, {\pname} introduces a unified
sequence processing approach that fundamentally differs from previous methods' rigid
feature addition. By directly concatenating condition tokens with noisy image
tokens in a unified sequence, we allow the multi-modal attention mechanism to discover
appropriate relationships between tokens—whether spatial or semantic—without imposing
artificial constraints. This flexibility is crucial for handling both spatially-aligned
tasks like edge-guided generation and non-aligned tasks like subject-driven
generation within a single framework.

Building on this unified approach, {\pname} incorporates a dynamic positioning strategy
that adaptively assigns position indices based on the task type, enabling true
omni-capability without task-specific architectural modifications. Furthermore,
for practical flexibility, we introduce an attention bias mechanism that allows users
to dynamically adjust the influence of image conditions at inference time,
providing crucial control over the generation process.

The effectiveness of our architecture depends on high-quality training data, particularly
for subject-driven generation. Recognizing that state-of-the-art DiT models can
generate remarkably consistent image pairs of the same subject, we develop an automated
data synthesis pipeline that creates Subjects200K—a dataset of over 200,000 diverse,
identity-consistent images that provides the rich training signals needed to
fully realize {\pname}'s potential.

In summary, we highlight our contributions as follows:
\begin{itemize}
    \item We propose {\pname}, a minimal and universal control framework for DiT
        models that requires only 0.1\% additional parameters while effectively handling
        both spatially-aligned and non-aligned tasks, demonstrating that
        extensive architectural modifications are unnecessary for effective
        image conditioning.

    \item We identify two key technical innovations that enable omni-capability:
        (1) unified sequence processing that outperforms traditional feature
        addition, and (2) adaptive position encoding that strategically assigns position
        indices based on task requirements.

    \item We design a flexible attention bias mechanism that allows precise adjustment
        of conditioning strength at inference time, enhancing practical control
        within the multi-modal framework without compromising performance.

    \item We develop and release Subjects200K, a large-scale dataset containing
        over 200,000 identity-consistent images to advance future research.
\end{itemize}

    \section{Related works}

\subsection{Diffusion models}

Diffusion models have emerged as a powerful framework for image generation\cite{ho2020denoising,
rombach2021highresolution}, demonstrating success across diverse tasks including
text-to-image synthesis~\cite{rombach2021highresolution, chen2023pixart,
saharia2022photorealistic}, image-to-image translation~\cite{saharia2022palette},
and image editing~\cite{meng2021sdedit, avrahami2022blended}. Recent advances have
led to significant improvements in both quality and efficiency, notably through
the introduction of latent diffusion models~\cite{rombach2021highresolution}. To
further enhance generative capabilities, large-scale transformer architectures
have been integrated into these frameworks, leading to advanced models like DiT\cite{peebles2023scalable,
chen2023pixart, blackforestlabs_flux, chen2024pixart}. Building on these architectural
innovations, FLUX\cite{blackforestlabs_flux} incorporates transformer-based design
with flow matching objectives~\cite{lipman2022flow}, achieving state-of-the-art
generation performance.

\subsection{Controllable generation}

Controllable generation has been extensively studied in the context of diffusion
models. Text-to-image models\cite{rombach2021highresolution, podell2023sdxl} have
established a foundation for conditional generation, while various approaches
have been developed to incorporate additional control signals such as image. Notable
methods include ControlNet~\cite{zhang2023adding}, enabling spatially aligned
control in diffusion models, and T2I-Adapter~\cite{mou2024t2i}, which improves
efficiency with lightweight adapters. UniControl~\cite{qin2023unicontrol} uses Mixture-of-Experts
(MoE) to unify different spatial conditions, further reducing model size. However,
these methods rely on spatially adding condition features to the denoising
network's hidden states, inherently limiting their effectiveness for spatially non-aligned
tasks like subject-driven generation. IP-Adapter~\cite{ye2023ip} addresses this
by introducing cross-attention through an additional encoder, and SSR-Encoder~\cite{zhang2024ssr}
further enhances identity preservation in image-conditioned tasks. Despite these
advances~\cite{li2024blip, ma2024subject, gal2022image, pan2023kosmos,ruiz2023dreambooth, hua2023dreamtunersingleimagesubjectdriven, kumari2023multi, he2025anystoryunifiedsinglemultiple, ma2024characteradapterpromptguidedregioncontrol}, a unified
solution for both spatially aligned and non-aligned tasks remains elusive.

    \section{Methods}
\subsection{ { Preliminary } }

The Diffusion Transformer (DiT) model~\cite{peebles2023scalable}, employed in
architectures like FLUX.1~\cite{blackforestlabs_flux}, Stable Diffusion 3~\cite{rombach2021highresolution},
and PixArt~\cite{chen2023pixart}, uses transformer as denoising network to
refine noisy image tokens iteratively.

\begin{figure*}[!t]
    \centering
    \begin{subfigure}
        [b]{0.8\textwidth}
        \centering
        \includegraphics[width=\textwidth]{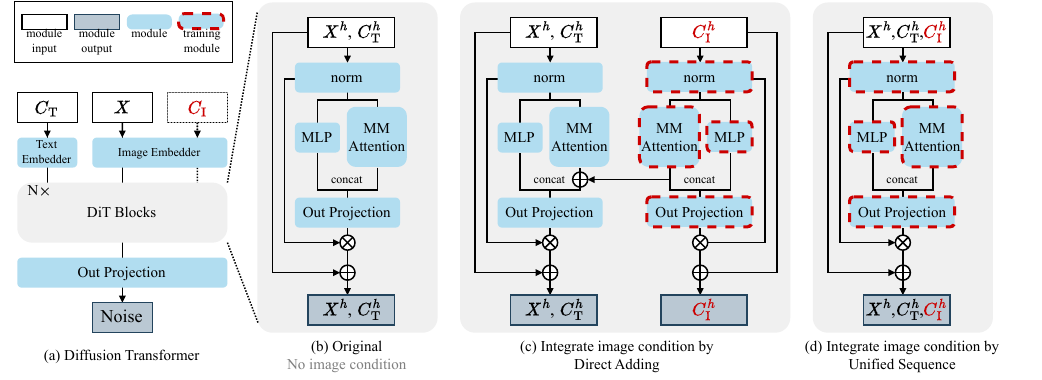}
        \caption{ Overview of the Diffusion Transformer (DiT) architecture and integration
        methods for image conditioning.}
        \label{fig:method}
    \end{subfigure}
    \hspace{-0.5cm}
    \begin{subfigure}
        [b]{0.18\textwidth}
        \centering
        \includegraphics[width=\textwidth]{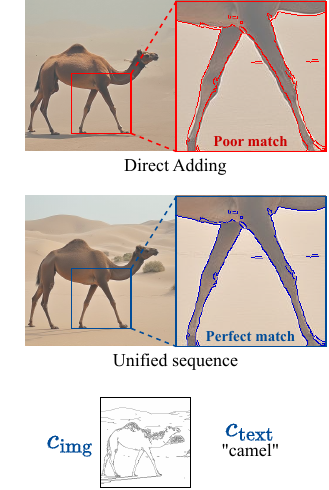}
        \caption{ Comparison of results }
        \label{fig:mmattn}
    \end{subfigure}
    \caption{Exploration of different methods for integrating image conditions.}
\end{figure*}

A DiT model processes two types of tokens: noisy image tokens ${X}\in \mathbb{R}^{N
\times d}$ and text condition tokens ${C}_{\text{T}}\in \mathbb{R}^{M \times d}$,
where $d$ is the embedding dimension, $N$ and $M$ are the number of image and text
tokens respectively (Figure~\ref{fig:method}). Throughout the network, these tokens
maintain consistent shapes as they pass through multiple transformer blocks.

In FLUX.1, each DiT block consists of layer normalization followed by {Multi-Modal Attention}
(MMA)~\cite{pan2020multi}, which incorporates {Rotary Position Embedding} (RoPE)~\cite{su2024roformer}
to encode spatial information. For image tokens ${X}$, RoPE applies rotation
matrices based on the token's position $(i,j)$ in the 2D grid:
\begin{equation}
    {X}_{i,j}\rightarrow{X}_{i,j}\cdot R(i,j),
\end{equation}
where $R(i,j)$ is the rotation matrix at position $(i,j)$. Text tokens
${C}_{\text{T}}$ undergo the same transformation with their positions set to
$(0,0)$.

The multi-modal attention mechanism then projects the position-encoded tokens
into query $Q$, key $K$, and value $V$ representations. It enables the computation
of attention between all tokens:
\begin{equation}
    \text{MMA}([{X};{C}_{\text{T}}]) = \text{softmax}\left(\frac{QK^{\top}}{\sqrt{d}}
    \right)V, \label{eq:mma}
\end{equation}
where $[{X};{C}_{\text{T}}]$ denotes the concatenation of image and text tokens.
This formulation enables bidirectional attention.


\subsection{\pname}

\begin{figure}[t]
    \centering
    \begin{subfigure}
        [b]{0.49\columnwidth}
        \centering
        \includegraphics[width=\linewidth, trim={7pt 7pt 6pt 7pt}, clip]{
            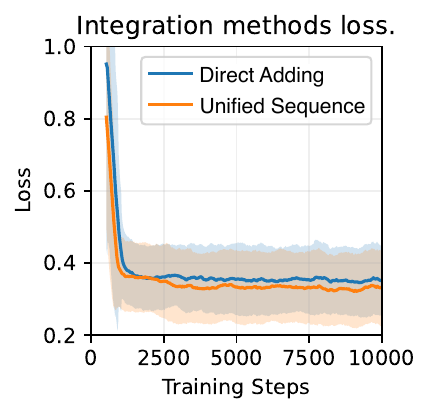
        }
        \caption{Training losses for different image condition integration
        methods.}
        \label{fig:pos_loss2}
    \end{subfigure}
    \hfill
    \begin{subfigure}
        [b]{0.47\columnwidth}
        \centering
        \includegraphics[width=\linewidth, trim={7pt 7pt 6pt 7pt}, clip]{
            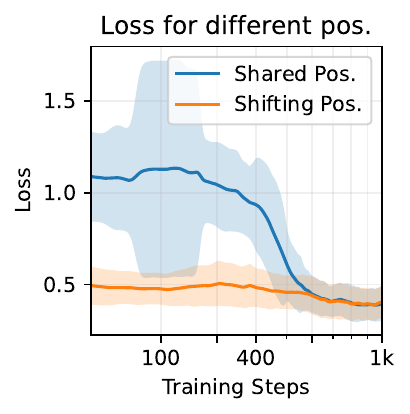
        }
        \caption{Training loss for shared vs. shifting position. (subject-driven)}
        \label{fig:pos_loss1}
    \end{subfigure}
    \caption{Training loss comparisons.}
    \label{fig:losses}
\end{figure}

Building upon the DiT architecture, with FLUX.1~\cite{blackforestlabs_flux} as our
base model, we aim to develop a \textbf{minimal}, \textbf{omni}-capable and \textbf{controllable}
generation framework that accepts flexible control signals. This vision leads to
our {\pname}, which we describe in this section.

\subsubsection{Minimal architecture}
To minimize the extra architectural and parameter overhead, {\pname} adopts a \emph{parameter
reuse strategy}.

As its name implies, we reuses the VAE~\cite{kingma2013auto, rombach2021highresolution}
encoder from the base DiT model encode the condition image. These images are
projected into the same latent space as the noisy input tokens, ensuring compatibility
without introducing new modules. Following this, {\pname} processes both noisy image
tokens and condition tokens jointly through the original DiT blocks. To adapt
the shared DiT blocks for handling condition tokens, only lightweight LoRA fine-tuning~\cite{devalal2018lora}
is applied, avoiding costly full-parameter updates.

This approach contrasts with previous methods that rely on separate feature
extractors like CLIP~\cite{radford2021learning, ye2023ip} or additional control
modules~\cite{zhang2024ssr}, significantly reducing architectural complexity.
Meanwhile, LoRA fine-tuning ensures parameter efficiency compared to duplicating
the entire network as done in ControlNet~\cite{zhang2023adding}.

\subsubsection{Omni-capable token interaction}
To achieve effective control across diverse tasks, {\pname} needs to enable
flexible interactions between condition tokens and image tokens. We address this
challenge through two key mechanisms: unified sequence processing and position-aware
token interaction.

\paragraph{Unified sequence processing.}

Building upon the shared latent space, we now think how to integrate condition tokens
into the model for flexible control across.


Previous methods~\cite{zhang2023adding, mou2024t2i} incorporate condition images
through direct feature adding:
\begin{equation}
    {h_X}\leftarrow{h_X}+h_{{C}_{\text{I}}},
\end{equation}
where the condition features $h_{{C}_{\text{I}}}$ are spatially aligned and added
to the noisy image features $h_{X}$.

We first implement this direct adding approach as illustrated in Figure~\ref{fig:method}.
While the bare effectiveness is shown in Figure~\ref{fig:mmattn}, this approach faces
two limitations: (1) it lacks flexibility for non-aligned scenarios where
spatial correspondence doesn't exist, and (2) the rigid addition operation constrains
potential interactions between condition and image tokens.

We then explored the unified sequence processing to integrate condition tokens into
the model. Specifically, this approach directly concatenates condition tokens with
noisy image tokens $[{X};{C}_{\text{T}};{C}_{\text{I}}]$ for multi-modal attention
processing. This formulation enables flexible token interactions through DiT's
multi-modal attention mechanism, allowing direct relationships to emerge between
any pair of tokens without imposing rigid spatial constraints.

\begin{figure}[t]
    \centering
    \includegraphics[width=\linewidth]{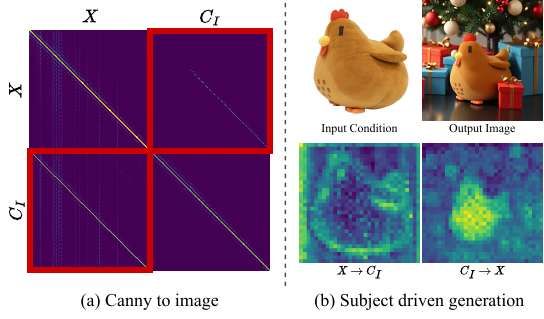}
    \caption{(a) Attention maps for the Canny-to-image task (with setting from Figure~\ref{fig:mmattn}),
    showing interactions between noisy image tokens $X$ and image condition tokens
    $C_{I}$. Strong diagonal patterns indicate effective spatial alignment. (b) Subject-driven
    generation task, with input condition and output image. (\textit{Prompt: This
    item is placed on a table with Christmas decorations around it.}) Attention
    maps for $X \to C_{i}$ and $C_{i}\to X$ illustrate accurate subject-focused
    attention.}
    \label{fig:attn}
\end{figure}

As shown in Figure~\ref{fig:attn}, this approach effectively handles both spatially
aligned and non-aligned tasks, with attention maps revealing clear cross-token
relationships and interaction patterns. Empirically, this unified sequence
approach consistently achieves lower training loss compared to direct feature
adding (Figure~\ref{fig:pos_loss2}), demonstrating its superior conditioning capability
across diverse generation scenarios.

\paragraph{Position-aware token interaction.}

While the unified sequence approach allows flexible token interactions, encoding
position information for the newly appended conditional tokens is not straightforward.


FLUX.1 adopts the RoPE mechanism to encode spatial information for both image and
text tokens. Specifically, for a 512$\times$512 input image, the VAE encoder
produces a 32$\times$32 grid of latent tokens, with each token assigned a position
index $(i,j)$ where $i ,j \in [0,31]$.

\begin{figure}[t]
    \centering
    \includegraphics[width=0.9\linewidth]{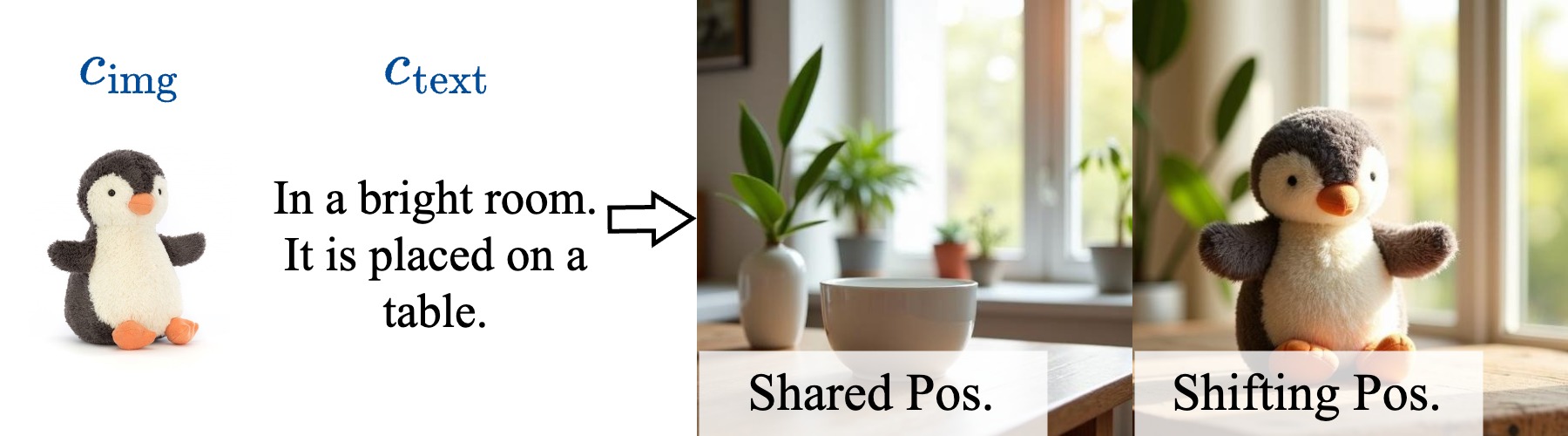}
    \caption{The results from models with shared and shifted position indices. Both
    models are fully trained with 15k iterations.}
    \label{fig:shifting}
\end{figure}

We first assign the same position indices to condition tokens as noisy image tokens,
it works well for spatially aligned tasks like edge-guided generation. However,
for non-aligned tasks like subject-driven generation, this shared position indexing
can lead to suboptimal performance due to spatial overlap between condition and noisy
image tokens. But if we shift the position indices of condition tokens by a
fixed offset $\Delta$ (e.g., $(0,32)$), ensuring no spatial overlap with noisy image
tokens, the training convergence accelerates significantly and final performance
improves. (see Figure~\ref{fig:pos_loss1} and Figure~\ref{fig:shifting}).

This observation suggests that for spatially aligned tasks, a shared position indexing
facilitates direct spatial correspondence between condition and image tokens,
while for non-aligned tasks, shared position indexing can constrain the model's ability
to establish semantic relationships. Hence, we propose a dynamic positioning strategy
based on the control task:
\begin{equation}
    (i,j)_{{C}_{\text{I}}}=
    \begin{cases}
        (i,j)_{{X}}    & \text{for aligned tasks}     \\
        (i,j) + \Delta & \text{for non-aligned tasks}
    \end{cases}
\end{equation}
which makes OminiControl truly omni-capable by adapting to both spatially
aligned and non-aligned control tasks.

\subsubsection{Control with flexibility}
\label{subsec:control}

Although our unified sequence processing and position-aware token interaction mechanisms
effectively enable joint attention between conditions and images, they also
present a new challenge. Unlike previous methods~\cite{ye2023ip, zhang2023adding}
that could simply scale condition features (e.g.,
$h_{X}\leftarrow h_{X}+ \alpha \cdot h_{C_{I}}$ where $\alpha$ controls strength),
the joint attention approach of {\pname} does not inherently support adjustable
conditioning strength, which is crucial for practical applications where users need
to balance text and image influences.

To address this limitation, we design a flexible control mechanism by
introducing a bias term into the multi-modal attention computation. Specifically,
for a given strength factor $\gamma$, we modify the attention operation in Equation~\ref{eq:mma}
to:
\begin{equation}
    \text{MMA}([{X};{C}_{\text{T}};{C}_{\text{I}}]) = \text{softmax}\left(\frac{QK^{\top}}{\sqrt{d}}
    + B(\gamma)\right)V,
\end{equation}
where $B(\gamma)$ is a bias matrix modulating the attention between concatenated
tokens $[{X};{C}_{\text{T}};{C}_{\text{I}}]$. Given ${C}_{\text{T}}\in \mathbb{R}
^{M \times d}$ and ${X},{C}_{\text{I}}\in \mathbb{R}^{N \times d}$, the bias
matrix has the structure:
\begin{equation}
    B(\gamma) =
    \begin{bmatrix}
        \mathbf{0}_{M \times M}             & \mathbf{0}_{M \times N} & \log(\gamma)\mathbf{1}_{N \times N} \\
        \mathbf{0}_{N \times M}             & \mathbf{0}_{N \times N} & \mathbf{0}_{N \times N}             \\
        \log(\gamma)\mathbf{1}_{N \times N} & \mathbf{0}_{M \times N} & \mathbf{0}_{N \times N}
    \end{bmatrix}.
\end{equation}
This formulation preserves the original attention patterns within each token type
while scaling attention weights between ${X}$ and ${C}_{\text{I}}$ by $\log(\gamma
)$. At test time, setting $\gamma = 0$ removes the condition's influence, while
$\gamma > 1$ enhances it, which makes OminiControl more flexible.

\subsubsection{Comparison with concurrent works}
Several recent works also explore controllable image generation with DiT models.
Some approaches~\cite{blackforestlabs_flux, mao2025ace++, han2024ace} employ
channel concatenation to integrate condition tokens, which offers less flexibility
than our unified sequence processing. Others~\cite{cai2024diffusion, choi2024style}
focus exclusively on specific tasks such as style transfer or subject-driven
generation, limiting their flexibility and generality. In contrast, our approach
combines dynamic positioning with a flexible control mechanism to provide a more
comprehensive and adaptable control framework compared to existing methods.

\begin{figure}[t]
    \centering
    \vspace{-1em}
    \includegraphics[width=\linewidth]{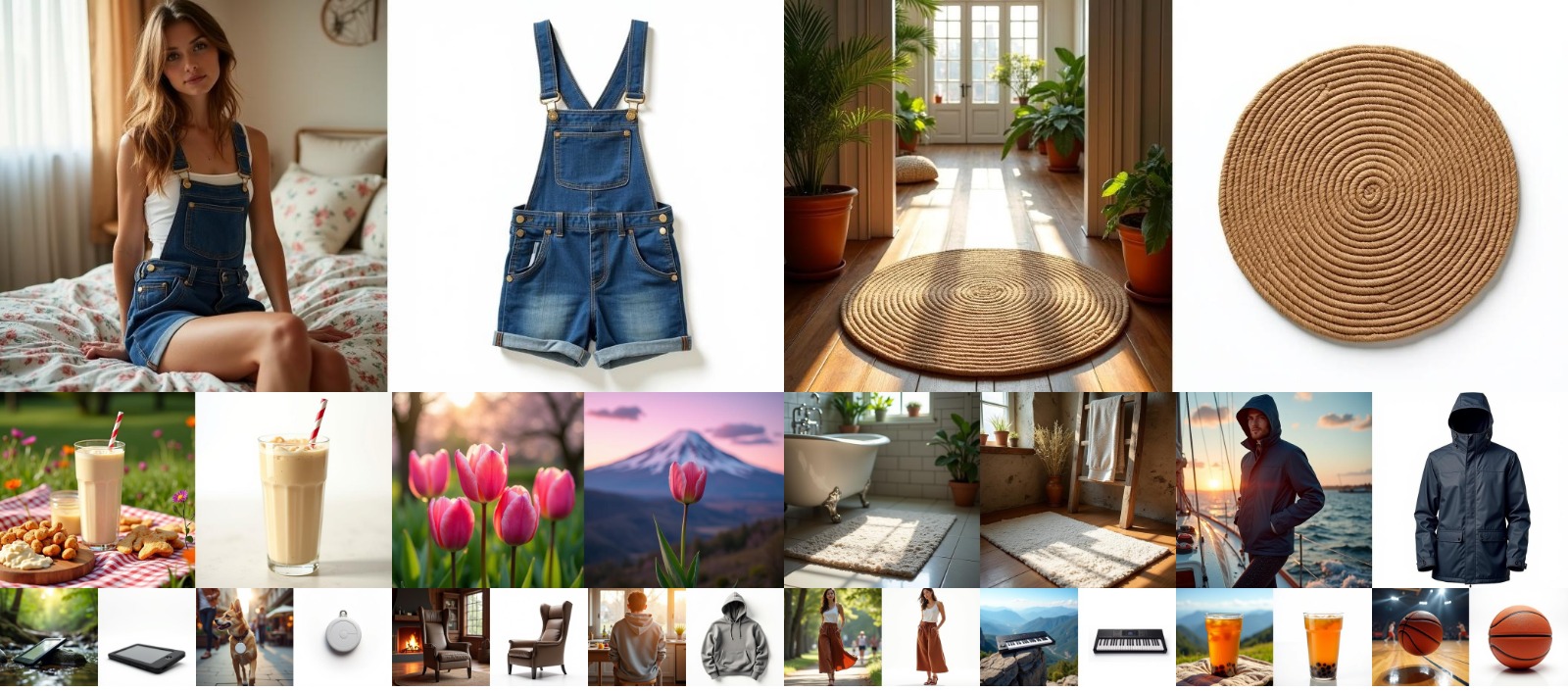}
    \caption{
    Examples from our \datasetname dataset. Each pair of images shows the same object
    in varying contexts.
    }
    \vspace{-0.5cm}
    \label{fig:dataset}
\end{figure}

\subsection{ { \datasetname\ datasets }}
\label{subsec:data}

A critical challenge in developing universal control frameworks lies in obtaining
high-quality training data for subject-driven generation. While \pname's
architecture enables flexible control, effective training requires data that maintains
subject consistency while incorporating natural variations in pose, lighting, and
context.

Existing solutions often rely on identical image pairs~\cite{ye2023ip} or limited-scale
datasets~\cite{ruiz2023dreambooth, kumari2023multi}, which present several
limitations. Using identical pairs can lead to overfitting in our framework,
causing the model to simply reproduce the input. Meanwhile, existing datasets
with natural variations often lack either the scale or diversity needed for
robust training.

To address this data challenge, we leverage a key observation: state-of-the-art
DiT models like FLUX.1~\cite{flux1controlnet2024} can generate identity-consistent
image pairs when provided with appropriate prompts~\cite{huang2024context, huang2024group}.
Building on this, we develop an pipeline to create Subjects200K, a large-scale
dataset specifically designed for subject-driven generation\footnote{Detailed
prompt design and synthesization pipeline are provided in the Section~\ref{ap:dataset}
of supplementary material.}:
\begin{itemize}
    \item \textbf{Prompt Generation}. We use GPT-4o to generate over 30,000 diverse
        subject descriptions. Each description represents the same subject in
        multiple scenes.


    \item \textbf{Paired-image synthesis}. We then reorganized the collected descriptions
        into structured prompts. Each prompt describes the same subject in two
        different scenes. The template of such prompt is shown in the Figure~\ref{lst:img_prompt}
        of supplementary material.These prompts are then fed into FLUX to generate
        image pairs. Each pair is designed to maintain subject consistency while
        varying in context.

    \item \textbf{Quality Assesment}. Finally, we use GPT-4o to evaluate the
        generated pairs. Misaligned pairs are removed to ensure identity consistency
        and high image quality.
\end{itemize}

The resulting dataset (Figure~\ref{fig:dataset}) contains over 200,000 high-quality
images spanning diverse categories. Each subject appears in multiple contexts, providing
rich training signals for learning robust subject-driven control. To facilitate
future research, we release both the dataset and our complete generation pipeline\footnote{Dataset
and code available at supplementary material.}.


    \section{Experiment}

\subsection{Setup}
\label{subsec:setup} \textbf{Tasks and base model.} We evaluate our method on
two categories of conditional generation tasks: spatially aligned tasks (including
Canny-to-image, depth-to-image, masked-based inpainting, and colorization) and
subject-driven generation. We build our method upon FLUX.1~\cite{blackforestlabs_flux},
a latent rectified flow transformer model for image generation. By default, we use
FLUX.1-dev to generate images for spatially aligned tasks. In subject-driven generation
tasks, we switch to FLUX.1-schnell as we observed it tend to produce better visual
quality.

\begin{figure*}[!t]
    \centering
    \includegraphics[width=\linewidth]{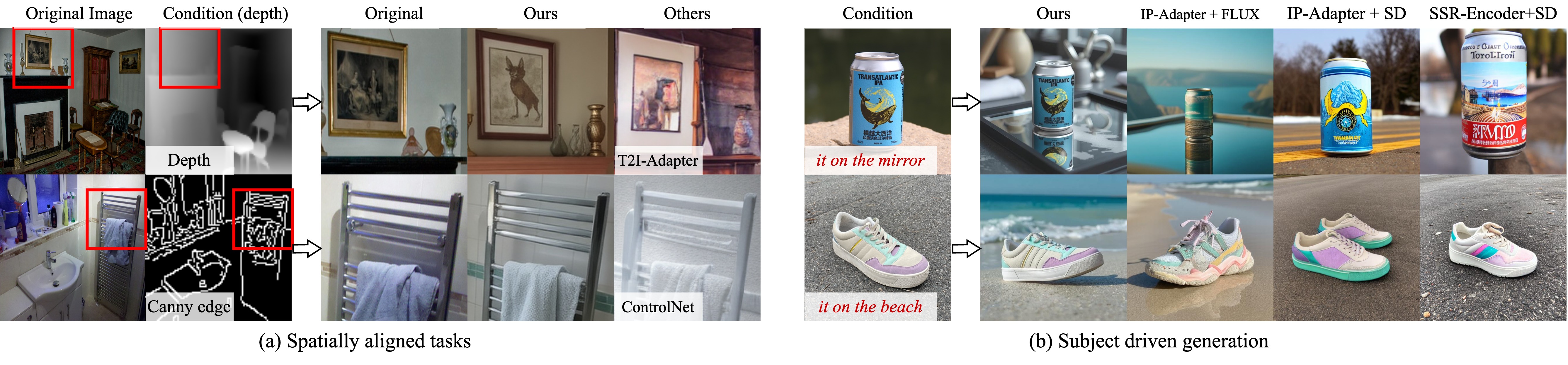}
    \caption{ Qualitative comparison. Left: Spatially aligned tasks. Right: Subject-driven
    generation with beverage can and shoes. Our method demonstrates superior
    controllability and quality across all tasks. (More results are provided in
    supplementary materials.) }
    \label{fig:all_comp}
\end{figure*}

\textbf{Implement details.} Our method utilizes LoRA~\cite{devalal2018lora}for
fine-tuning the base model with a default rank of 4. To preserve the model's
original capabilities and achieve flexibility, the LoRA scale is set to 0 by default when processing non-condition tokens.

\textbf{Training.} Our model is trained with a batch size of 1 and gradient
accumulation over 8 steps (effective batch size of 8). We employ the Prodigy optimizer~\cite{mishchenko2024prodigy}
with safeguard warmup and bias correction enabled, setting the weight decay to
0.01. For spatially aligned tasks, we use text-to-image-2M\cite{jackyhate2024t2i}
dataset with the last 300,000 images. For subject-driven generation, we utilize our
proposed Subjects200K dataset. The experiments are conducted on 2 NVIDIA H100
GPUs (80GB each). For spatially aligned tasks, models are trained for 50,000
iterations, while subject-driven generation models are trained for 15,000 iterations.

\textbf{Baselines.} For spatially aligned tasks, we compare our method with both
the original ControlNet~\cite{zhang2023adding} and T2I-Adapter~\cite{mou2024t2i}
on Stable Diffusion 1.5, as well as ControlNetPro~\cite{flux1controlnet2024}, the
FLUX.1 implementation of ControlNet. For subject-driven generation, we compare
with IP-Adapter~\cite{ye2023ip}, evaluating its implementations FLUX.1~\cite{fluxipadapter2024}.
Additionally, we also with the official FLUX.1 Tools~\cite{blackforestlabs_flux}
implementation.

\textbf{Evaluation of spatially aligned tasks.} We evaluate our model on both
spatially aligned tasks and subject-driven generation. For spatially aligned tasks,
we assess two aspects: generation quality and controllability. Generation
quality is measured using FID~\cite{heusel2017gans}, SSIM, CLIP-IQA\cite{wang2022exploringclipassessinglook},
MAN-IQA~\cite{yang2022maniqa}, MUSIQ~\cite{ke2021musiq}, and PSNR\cite{Wikipedia2024psnr}
for visual fidelity, along with CLIP Text and CLIP Image scores~\cite{radford2021learning}
for consistency. For controllability, we compute F1 Score between extracted and
input edge maps in edge-conditioned generation, and MSE between extracted and original
condition maps for other tasks (using Depth Anything for depth, color channel separation
for colorization, etc.). We use 5,000 images from COCO 2017 validation set, and
resize them to 512$\times$512, then generate task-specific conditions and associated
captions as prompts with a fixed seed of 42.

\textbf{Evaluation of subject-driven generation.} For subject-driven generation,
we propose a five-criteria framework evaluating both preservation of subject
characteristics (identity preservation, material quality, color fidelity,
natural appearance) and accuracy of requested modifications, with all
assessments conducted through GPT-4o's vision capabilities to ensure systematic
evaluation. Details are presented in the Section~\ref{subsec:ap:sub} of supplementary
material. We test on 750 text-condition pairs (30 subjects × 25 prompts) from
DreamBooth\cite{ruiz2023dreambooth} dataset with 5 different seeds, using one
selected image per subject as the condition.

\begin{figure}[t]
    \centering
    \vspace{-1em}
    \includegraphics[width=\linewidth]{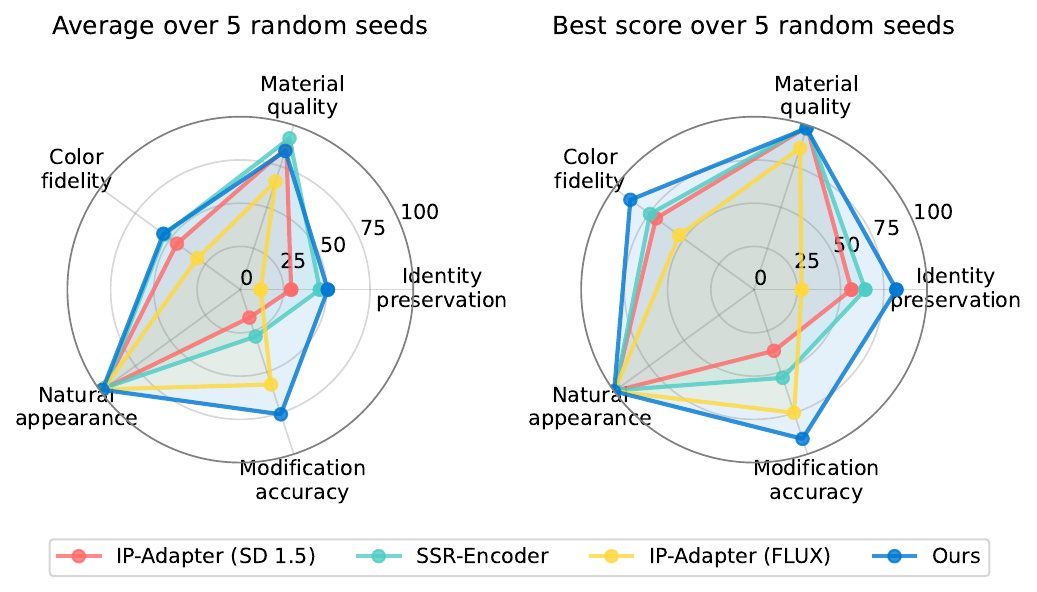}
    \caption{
    Radar charts visualization comparing our method (blue) with baselines across
    five evaluation metrics. }
    \vspace{-1em}
    \label{fig:radar}
\end{figure}

\subsection{Main result}

\textbf{Spatially aligned tasks} As shown in Table~\ref{tab:main_results}, we
comprehensively evaluate our method against existing approaches on five
spatially aligned tasks. Our method achieves the highest F1-Score of 0.38 on depth-to-image
generation, significantly outperforming both SD1.5-based methods ControlNet~\cite{zhang2023adding}
and T2I-Adapter~\cite{mou2024t2i}, as well as FLUX.1-based ControlNetPro~\cite{flux1controlnet2024}.
In terms of general quality metrics, our approach demonstrates consistent superiority
across most tasks, showing notably better performance in SSIM~\cite{wang2004image},
CLIP-IQA\cite{wang2022exploringclipassessinglook}, MAN-IQA~\cite{yang2022maniqa},
MUSIQ~\cite{ke2021musiq} and PSNR\cite{Wikipedia2024psnr} scores. For
challenging tasks like deblurring and colorization, our method achieves substantial
improvements: the MSE is reduced by 77\% and 93\% respectively compared to
ControlNetPro, while the FID scores~\cite{heusel2017gans} improve from 30.38 to 11.49
for deblurring. The CLIP-Text and CLIP-Image metrics~\cite{radford2021learning}
indicate that our method maintains high consistency across all tasks, suggesting
effective preservation of semantic alignment and image alignment while achieving
better control and visual quality. As shown in Figure~\ref{fig:all_comp}, our method
produces sharper details and more faithful color reproduction in colorization tasks,
while maintaining better structural fidelity in edge-guided generation and
deblurring scenarios.

\begin{table}[t]
    \scriptsize
    \centering
    \begin{tabular}{l|c|c|c}
        \toprule {Methods}                   & {Base model}                                               & {Parameters}                                            & {Ratio}                                                             \\
        \midrule \midrule ControlNet         & \multirow{3}{*}{\makecell{SD1.5 \textcolor{gray}{/} 860M}} & 361M                                                    & $\sim$42\%                                                          \\
        T2I-Adapter                          &                                                            & 77M                                                     & $\sim$9.0\%                                                         \\
        IP-Adapter                           &                                                            & 449M                                                    & $\sim$52.2\%                                                        \\
        \midrule ControlNet                  & \multirow{3}{*}{\makecell{FLUX.1 \textcolor{gray}{/} 12B}} & 3.3B                                                    & $\sim$27.5\%                                                        \\
        IP-Adapter                           &                                                            & 918M                                                    & $\sim$7.6\%                                                         \\
        FLUX.1 Tools                         &                                                            & 612M                                                    & $\sim$5.1\%                                                         \\
        \midrule \rowcolor{cvprblue!15} Ours & FLUX.1 \textcolor{gray}{/} 12B                             & \makecell{14.5M \textcolor{gray}{/} \\48.7M w/ Encoder} & \makecell{$\sim$0.1\% \textcolor{gray}{/} \\$\sim$0.4\% w/ Encoder} \\
        \bottomrule
    \end{tabular}
    \caption{Additional parameters introduced by different image conditioning
    methods. For IP-Adapter, the parameter count includes the CLIP Image encoder.
    For our method, we also report results when using the original VAE encoder
    from FLUX.1.}
    \vspace{-1em}
    \label{table:param}
\end{table}

\newcommand{\best}[1]{\textbf{#1}}

\begin{table*}
    [t] \scriptsize
    \centering
    \resizebox{\textwidth}{!}{
    \begin{tabular}{c|c|c|c|cccccc|cc}
        \toprule \multirow{3}{*}{Task}           & \multirow{3}{*}{Methods / Setting} & \multirow{3}{*}{ Base Model} & Controllability                                & \multicolumn{6}{c|}{Image Quality}  & \multicolumn{2}{c}{Alignment}       \\
                                                 &                                    &                              & \textbf{F1}$\uparrow$ / \textbf{MSE}$\downarrow$ & \textbf{FID}$\downarrow$            & \textbf{SSIM} $\uparrow$           & \textbf{CLIP-IQA}$\uparrow$         & \textbf{MAN-IQA}$\uparrow$         & \textbf{MUSICQ}$\uparrow$                & \textbf{PSNR} $\uparrow$                 & \textbf{CLIP Text}$\uparrow$             & \textbf{CLIP Image}$\uparrow$       \\
        \midrule \midrule \multirow{5}{*}{Canny} & ControlNet                         & \multirow{2}{*}{SD1.5}       & 0.35                                           & \best{18.74}                        & \underline{0.36}                   & 0.65                                & 0.45                               & 67.81                                    & \underline{10.27}                        & 0.305                                    & \underline{0.752}                   \\
                                                 & T2I Adapter                        &                              & 0.22                                           & \underline{20.06}                   & 0.35                               & 0.57                                & 0.39                               & 67.89                                    & 9.53                                     & 0.305                                    & 0.748                               \\
                                                 & ControlNet Pro                     & \multirow{3}{*}{FLUX.1}      & 0.21                                           & 98.69                               & 0.25                               & 0.48                                & 0.37                               & 56.91                                    & 9.22                                     & 0.192                                    & 0.537                               \\
                                                 & Flux Tools                         &                              & 0.20                                           & 22.13                               & 0.32                               & \underline{0.66}                    & \underline{0.60}                   & \best{75.47}                             & 9.69                                     & \best{0.308}                             & 0.701                               \\
                                                 & Ours                               &                              & \best{0.50}\cellcolor{cvprblue!15}             & 24.20\cellcolor{cvprblue!15}        & \best{0.45}\cellcolor{cvprblue!15} & \best{0.66}\cellcolor{cvprblue!15}  & \best{0.62}\cellcolor{cvprblue!15} & \underline{74.87}\cellcolor{cvprblue!15} & \best{11.34}\cellcolor{cvprblue!15}      & \underline{0.305}\cellcolor{cvprblue!15} & \best{0.785}\cellcolor{cvprblue!15} \\
        \midrule \multirow{5}{*}{Depth}          & ControlNet                         & \multirow{2}{*}{SD1.5}       & 923                                            & \best{23.03}                        & \underline{0.34}                   & 0.64                                & 0.47                               & 70.73                                    & \best{10.63}                             & 0.308                                    & \underline{0.726}                   \\
                                                 & T2I Adapter                        &                              & 1560                                           & 24.72                               & 0.28                               & 0.61                                & 0.39                               & 69.99                                    & 9.50                                     & \best{0.309}                             & 0.721                               \\
                                                 & ControlNet Pro                     & \multirow{3}{*}{FLUX.1}      & 2958                                           & 62.20                               & 0.26                               & 0.55                                & 0.39                               & 66.85                                    & 9.38                                     & 0.212                                    & 0.547                               \\
                                                 & Flux Tools                         &                              & \underline{767}                                & \underline{24.56}                   & 0.32                               & \underline{0.68}                    & \underline{0.59}                   & \best{75.30}                             & 10.15                                    & \underline{0.308}                        & 0.715                               \\
                                                 & Ours                               &                              & \best{537}\cellcolor{cvprblue!15}              & 31.04\cellcolor{cvprblue!15}        & \best{0.39}\cellcolor{cvprblue!15} & \best{0.68}\cellcolor{cvprblue!15}  & \best{0.60}\cellcolor{cvprblue!15} & \underline{74.04}\cellcolor{cvprblue!15} & \underline{10.53}\cellcolor{cvprblue!15} & 0.305\cellcolor{cvprblue!15}             & \best{0.749}\cellcolor{cvprblue!15} \\
        \midrule \multirow{3}{*}{Mask}           & ControlNet                         & SD1.5                        & 7588                                           & 13.14                               & 0.68                               & \underline{0.58}                    & 0.42                               & 67.22                                    & \underline{18.96}                        & 0.300                                    & 0.848                               \\
                                                 & Flux Tools                         & \multirow{2}{*}{FLUX.1}      & \underline{6610}                               & \underline{11.40}                   & \underline{0.73}                   & 0.56                                & \underline{0.45}                   & \underline{68.92}                        & 18.37                                    & \underline{0.305}                        & \underline{0.874}                   \\
                                                 & Ours                               &                              & \best{6351}\cellcolor{cvprblue!15}             & \best{10.20}\cellcolor{cvprblue!15} & \best{0.78}\cellcolor{cvprblue!15} & \best{0.59 }\cellcolor{cvprblue!15} & \best{0.49}\cellcolor{cvprblue!15} & \best{70.78}\cellcolor{cvprblue!15}      & \best{19.59}\cellcolor{cvprblue!15}      & \best{0.305}\cellcolor{cvprblue!15}      & \best{0.892}\cellcolor{cvprblue!15} \\
        \midrule \multirow{2}{*}{Colorization}   & ControlNet Pro                     & \multirow{2}{*}{FLUX.1}      & 994                                            & 30.38                               & 0.75                               & 0.40                                & 0.31                               & 54.38                                    & 16.23                                    & 0.279                                    & 0.781                               \\
                                                 & Ours                               &                              & \best{73}\cellcolor{cvprblue!15}               & \best{10.37}\cellcolor{cvprblue!15} & \best{0.92}\cellcolor{cvprblue!15} & \best{0.56}\cellcolor{cvprblue!15}  & \best{0.48}\cellcolor{cvprblue!15} & \best{69.40}\cellcolor{cvprblue!15}      & \best{21.56}\cellcolor{cvprblue!15}      & \best{0.305}\cellcolor{cvprblue!15}      & \best{0.884}\cellcolor{cvprblue!15} \\
        \midrule \multirow{2}{*}{Deblur}         & ControlNet Pro                     & \multirow{2}{*}{FLUX.1}      & 338                                            & \best{16.27}                        & \best{0.64}                        & 0.55                                & 0.43                               & 70.95                                    & 20.53                                    & 0.294                                    & 0.853                               \\
                                                 & Ours                               &                              & \best{62}\cellcolor{cvprblue!15}               & 18.89\cellcolor{cvprblue!15}        & 0.58\cellcolor{cvprblue!15}        & \best{0.59 }\cellcolor{cvprblue!15} & \best{0.54}\cellcolor{cvprblue!15} & \best{70.98}\cellcolor{cvprblue!15}      & \best{21.82}\cellcolor{cvprblue!15}      & \best{0.301}\cellcolor{cvprblue!15}      & \best{0.856}\cellcolor{cvprblue!15} \\
        \midrule
    \end{tabular}
    }
    \caption{Quantitative comparison with baseline methods on five spatially
    aligned tasks. We evaluate methods based on Controllability (F1-Score for
    Canny, MSE for others), Image Quality (SSIM, FID, CLIP-IQA, MAN-IQA, MUSIQ, PSNR),
    and Alignment (CLIP Text and CLIP Image). For F1-Score (used in Canny to Image
    task), higher is better; for MSE, lower is better. Best results are shown in
    \best{bold}.The second-best results are highlighted with underlines.}
    \vspace{-1em}
    \label{tab:main_results}
\end{table*}

\textbf{Subject driven generation} Figure \ref{fig:radar} presents a comprehensive
comparison against existing baselines. Our method demonstrates superior
performance, particularly in identity preservation and modification accuracy.
Averaging over random seeds, we achieve 75.8\% modification accuracy compared to
IP-Adapter (FLUX)'s 57.7\%, while maintaining 50.6\% identity preservation against
IP-Adapter (SD 1.5)'s 29.4\%. The advantage amplifies in best-seed scenarios,
achieving 90.7\% modification accuracy and 82.3\% identity preservation -
surpassing the strongest baselines by 15.8 and 18.0 percentage points, demonstrating
effective subject-fidelity editing. These quantitative results are further
corroborated by user studies presented in supplementary material~\ref{subsec:ap:sub}.

\textbf{Paramter efficiency} As shown in Table~\ref{table:param}, our approach
achieves remarkable parameter efficiency compared to existing methods. For the 12B
parameter FLUX.1 model, our method requires only 14.5M trainable parameters (approximately
0.1\%), which is significantly lower than ControlNet (27.5\%) and IP-Adapter (7.6\%).
Even when utilizing the original VAE encoder from FLUX.1, our method still maintains
high efficiency with just 0.4\% additional parameters, demonstrating the effectiveness
of our parameter-efficient design.

\begin{figure}[t]
    \centering
    \vspace{-1em}
    \includegraphics[width=\linewidth]{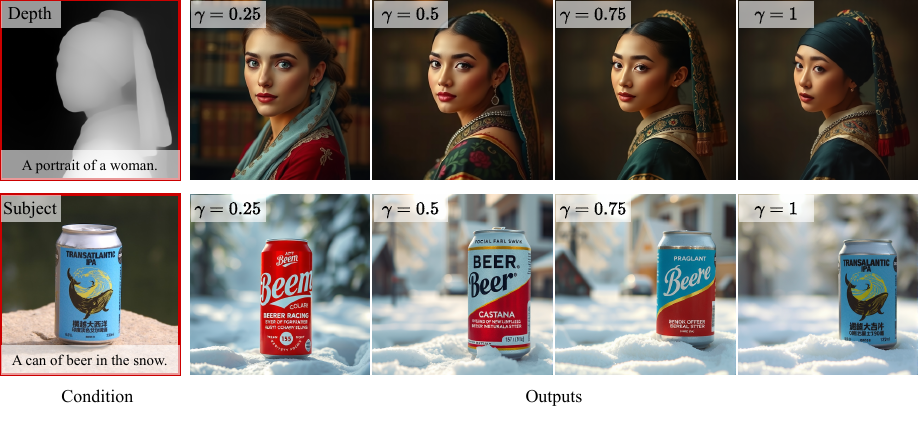}
    \vspace{-2em}
    \caption{Demonstration of the condition strength control.}
    \vspace{-1em}
    \label{fig:factor}
\end{figure}

\textbf{Condition strength factor} We evaluate our condition strength control
mechanism (Section~\ref{subsec:control}) through qualitative experiments. Figure~\ref{fig:factor}
shows generated results with varying strength factor $\gamma$. Results show that
$\gamma$ effectively controls the generation process for both spatially aligned
tasks like depth-to-image generation and non-aligned tasks like subject-driven
generation, enabling flexible control over the condition's influence.

\subsection{Ablation Studies}
To better understand the key factors that influence our model's control capabilities,
we conducted comprehensive ablation studies examining parameter efficiency, architectural
decisions, and component contributions.

\textbf{Impact of LoRA rank.} We conducted extensive experiments with different LoRA
ranks (1, 2, 4, 8, and 16) for the Canny-to-image task. As shown in Table~\ref{tab:ablation_studies},
our experiments show that increasing the LoRA rank generally improves model performance,
with rank 16 achieving the best results across multiple aspects. However, even
with smaller ranks (e.g., rank 1), the model demonstrates competitive performance,
especially in text-image alignment, showing the efficiency of our approach even with
limited parameters.

\begin{table}[t]
    \centering
    \scriptsize
    \begin{tabular}{c|c|cccc}
        \toprule Study                                           & Setting                      & FID $\downarrow$                       & SSIM $\uparrow$                        & F1 Score $\uparrow$                   & CLIP Score $\uparrow$                 \\
        \midrule \midrule \multirow{5}{*}{\makecell{LoRA\\Rank}} & 1                            & 21.09                                  & 0.412                                  & 0.385                                 & \textbf{0.765}                        \\
                                                                 & 2                            & 21.28                                  & 0.411                                  & 0.377                                 & 0.751                                 \\
                                                                 & \cellcolor{cvprblue!15} 4    & \cellcolor{cvprblue!15} 20.63          & \cellcolor{cvprblue!15} 0.407          & \cellcolor{cvprblue!15} 0.380         & \cellcolor{cvprblue!15} 0.761         \\
                                                                 & 8                            & 21.40                                  & 0.404                                  & 0.3881                                & 0.761                                 \\
                                                                 & 16                           & \textbf{19.71}                         & \textbf{0.425}                         & \textbf{0.407}                        & 0.764                                 \\
        \midrule \multirow{2}{*}{\makecell{Condition\\Blocks}}   & Early                        & 25.66                                  & 0.369                                  & 0.23                                  & 0.72                                  \\
                                                                 & \cellcolor{cvprblue!15} Full & \cellcolor{cvprblue!15} \textbf{20.63} & \cellcolor{cvprblue!15} \textbf{0.407} & \cellcolor{cvprblue!15} \textbf{0.38} & \cellcolor{cvprblue!15} \textbf{0.76} \\
        \bottomrule
    \end{tabular}
    \caption{Ablation studies on (1) LoRA rank for the \textit{Canny-to-image} task
    and (2) condition signal integration approaches. Results show that LoRA rank
    of 16 and full-depth integration achieve the best performance. Rows with blue
    background indicate our default settings (LoRA rank=4, Full condition integration).
    Best results are in \textbf{bold}.}
    \vspace{-1em}
    \label{tab:ablation_studies}
\end{table}

\textbf{Conditioning depth.} FLUX.1's transformer architecture features two
distinct types of blocks: early blocks that employ separate normalization modules
for different modalities tokens (text and image) and later blocks that share unified
normalization across all tokens. As shown in Table~\ref{tab:ablation_studies}, experiments
reveal that restricting condition signal integration to only these early blocks results
in insufficient controllability. This suggests that allowing the condition
signals to influence the entire transformer stack is crucial for generate better
results. Notably, this finding indicates that the preview approaches\cite{flux1controlnet2024,
fluxipadapter2024, ye2023ip, zhang2023adding, mou2024t2i} of injecting condition
signals primarily in early blocks, which were effective in UNet-based
architectures, may not fully translate to DiT-based models like FLUX.1.

\begin{figure}[t]
    \centering
    \includegraphics[width=\linewidth]{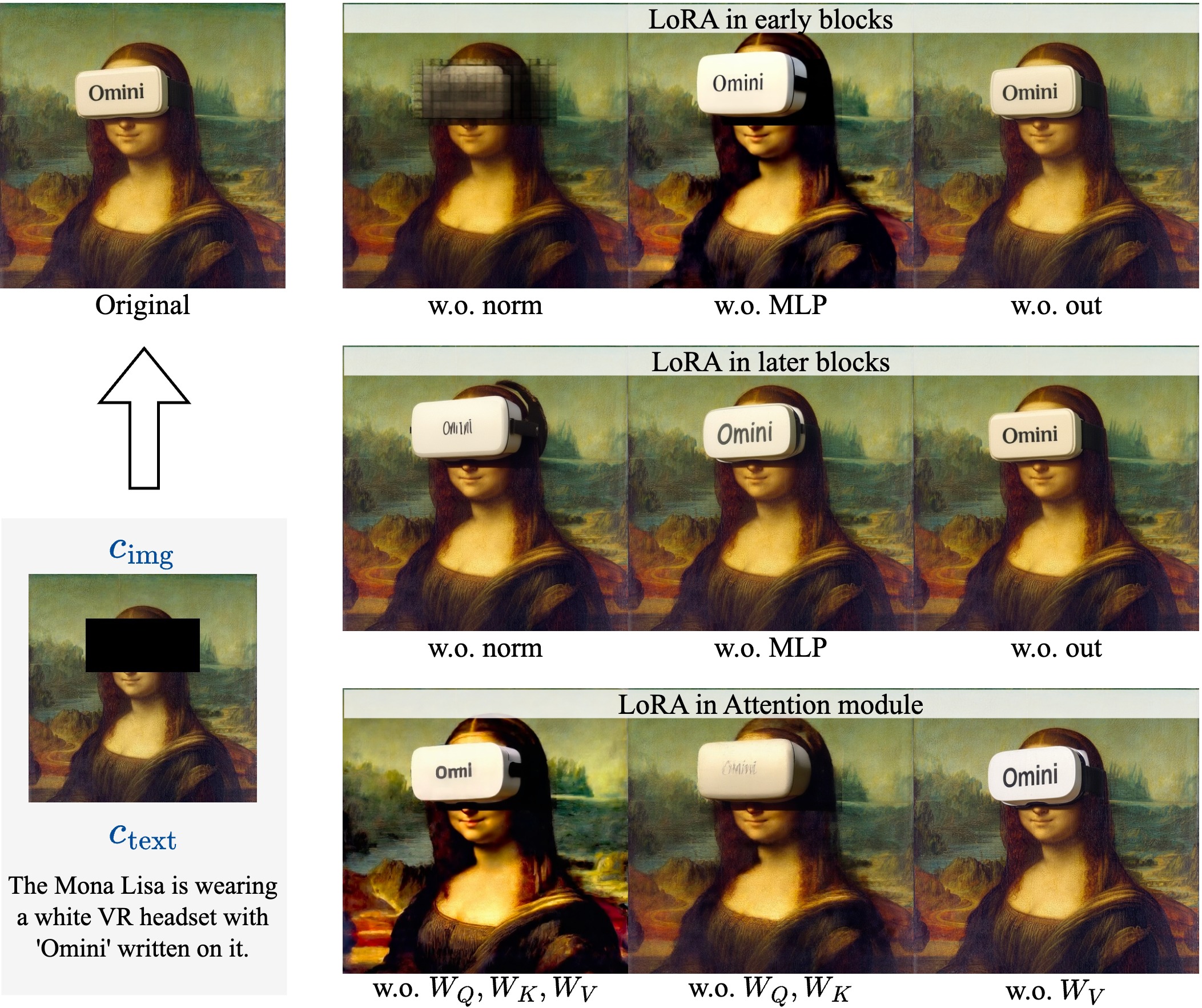}
    \caption{Ablation study of critical modules for conditional generation. Given
    the censored Mona Lisa image and text prompt (bottom left), we test removing
    LoRA from different components. }
    \vspace{-1em}
    \label{fig:lora_ablation}
\end{figure}

\textbf{Critical module analysis.} To identify essential components for
effective control, we conducted fine-grained ablation studies as shown in Figure~\ref{fig:lora_ablation}.
Results demonstrate that normalization layers and attention
projections—particularly query ($W_{Q}$) and key ($W_{K}$)—are critical for maintaining
control quality. Removing LoRA from these components significantly degrades
conditional rendering, while value projections ($W_{V}$) have less impact. These
insights reveal that control signals might propagate through normalization
pathways and attention routing mechanisms rather than feature transformation
processes.
    \section{Conclusion}
\pname~offers parameter-efficient image-conditional control for DiT across
diverse tasks using a unified token approach, requiring only 0.1\% additional
parameters. The Subjects200K dataset — featuring over 200,000 high-quality,
subject-consistent images—further supports advancements in subject-driven
generation, with experimental results confirming \pname's effectiveness in both spatially-aligned
and non-aligned tasks.

However, the unified sequence approach increases the total number of tokens
processed through the network, potentially limiting computational efficiency during
inference. Addressing this token efficiency challenge while maintaining our
method's control capabilities represents an important direction for future
research in parameter-efficient conditional generation.

    \section{Acknowledgment}
We would like to acknowledge that the computational work involved in this research work is partially supported by NUS IT’s Research Computing group using grant numbers NUSREC-HPC-00001.

    { \small \bibliographystyle{ieeenat_fullname} \bibliography{main} }
    \appendix

    \clearpage
    \setcounter{page}{1}
    \setcounter{figure}{0}
    \renewcommand{\thefigure}{S\arabic{figure}}
    \setcounter{table}{0}
    \renewcommand{\thetable}{S\arabic{table}}
    \maketitlesupplementary

    \section{Details of \datasetname\ datasets}
\label{ap:dataset} We present a comprehensive synthetic dataset constructed to
address the limitations in scale and image quality found in previous datasets~\cite{ruiz2023dreambooth,
kumari2023multi, li2024blip, li2024photomaker}. Our approach leverages FLUX.1-dev~\cite{blackforestlabs_flux}
to generate high-quality, consistent images of the same subject under various conditions.

\datasetname\ dataset currently consists of two splits, both generated using
similar pipelines. Split-1 contains paired images of objects in different scenes,
while Split-2 pairs each object's scene image with its corresponding studio
photograph. Due to their methodological similarities, we primarily focus on
describing the synthesis process and details of Split-2, although both splits are
publicly available.
{
Our complete Subjects200K dataset can be fully accessed via this \href{https://github.com/Yuanshi9815/Subjects200K}{link}. }

\subsection{Generation pipeline}
Our dataset generation process consists of three main stages: description
generation, image synthesis, and quality assessment.

\textbf{Description Generation} We employed ChatGPT-4o to create a hierarchical structure
of descriptions: We first generated 42 diverse object categories, including furniture,
vehicles, electronics, clothing, and others. For each category, we created multiple
object instances, totaling 4,696 unique objects. Each object entry consists of:
(1) A brief description, (2) Eight diverse scene descriptions, (3) One studio
photo description. Figure~\ref{lst:example} shows a representative example of our
structured description format.

\textbf{Image Synthesis} We designed a prompt template to leverage FLUX's
capability of generating paired images containing the same subject. Our template
synthesizes a comprehensive prompt by combining a brief object description with two
distinct scene descriptions, ensuring subject consistency while introducing
environmental variations.

The detailed prompt structure is illustrated in Figure~\ref{lst:img_prompt}. For
each prompt, we set the image dimensions to 1056×528 pixels and generated five
images using different random seeds to ensure diversity in our dataset. During the
training process, we first split the paired images horizontally, then performed central
cropping to obtain 512×512 pixel image pairs. This padding strategy was
implemented to address cases where the generated images were not precisely
bisected, preventing potential artifacts from appearing in the wrong half of the
split images.

\begin{figure}[h]
    \centering
    \includegraphics[width=\linewidth]{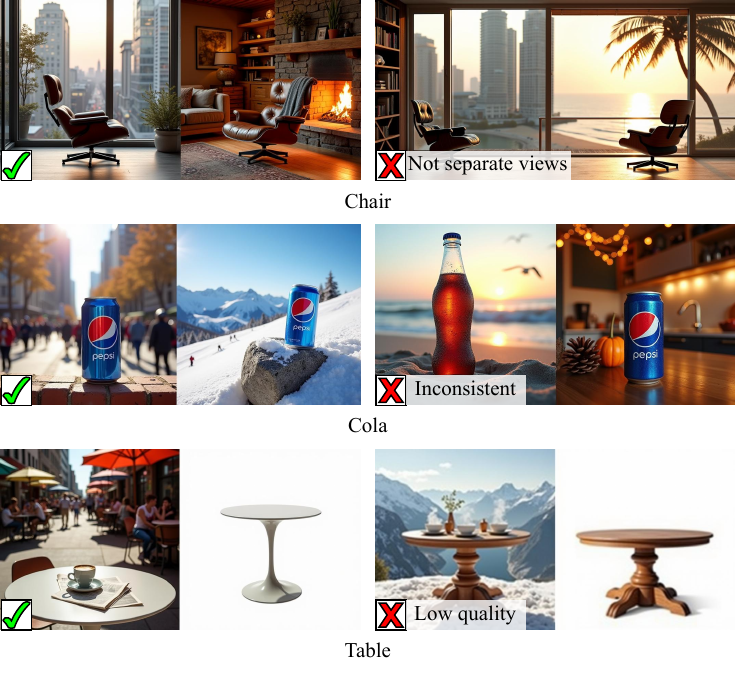}
    \caption{
    Examples of successful and failed generation results from \datasetname\ dataset.
    Green checks indicate successful cases where subject identity and characteristics
    are well preserved, while red crosses show failure cases. }
    \label{fig:data_cases}
\end{figure}

\textbf{Quality assessment} We leveraged ChatGPT-4o's vision capabilities to rigorously
evaluate the quality of images generated by FLUX.1-dev. The assessment focused on
multiple critical aspects:
\begin{itemize}
    \item Image composition: Verifying that each image properly contains two
        side-by-side views.

    \item Subject consistency: Ensuring the subject maintains identity across
        both views.

    \item Image quality: Confirming high resolution and visual fidelity.
\end{itemize}
To maintain stringent quality standards, each image underwent five independent evaluations
by ChatGPT-4o. Only images that passed all five evaluations were included in our
training dataset. Figure~\ref{fig:data_cases} presents representative examples from
our quality-controlled dataset.

\begin{figure*}[t]
    \centering
    \definecolor{backgroundcolor}{RGB}{250,250,250}
    \begin{minipage}{\textwidth}
        \lstset{ language=Python, basicstyle=\small\ttfamily, backgroundcolor=
        \color{backgroundcolor}
        , xleftmargin=0pt, breaklines=true, frame=single, showstringspaces=false,
        keepspaces=true } \begin{lstlisting}
{
    "brief_description": 
        "A finely-crafted wooden seating piece.",
    "scene_descriptions": [
        "Set on a sandy shore at dusk, it faces the ocean with a gentle breeze rustling nearby palms, bathed in soft, warm twilight.",
        "Positioned in a bustling urban cafe, it stands out against exposed brick walls, capturing the midday sun through a wide bay window."
        // Additional six scene descriptions omitted
    ],
    "studio_photo_description":
        "In a professional studio against a plain white backdrop, it is captured in three-quarter view under uniform high-key lighting, showcasing the delicate grain and smooth of its finely-crafted surfaces."
}
\end{lstlisting}
        \caption{An example of our structured description format for dataset
        generation.}
        \label{lst:example}
    \end{minipage}
\end{figure*}

\begin{figure*}[!t]
    \centering
    \definecolor{backgroundcolor}{RGB}{250,250,250}
    \begin{minipage}{\textwidth}
        \lstset{ language=Python, basicstyle=\small\ttfamily, backgroundcolor=
        \color{backgroundcolor}
        , xleftmargin=0pt, breaklines=true, frame=single, showstringspaces=false,
        keepspaces=true } \begin{lstlisting}
prompt_1 = f"Two side-by-side images of the same object: {brief_description}"
prompt_2 = f"Left: {scene_description1}"
prompt_3 = f"Right: {scene_description2}"
prompt_image = f"{prompt_1}; {prompt_2}; {prompt_3}"
\end{lstlisting}
        \caption{Our prompt template for paired image generation. The template combines
        a brief object description with two distinct scene descriptions to maintain
        subject consistency while varying environmental conditions.}
        \label{lst:img_prompt}
    \end{minipage}
\end{figure*}

\subsection{Dataset Statistics}
In Split-2, we first generated 42 distinct object categories, from which we created
and curated a set of 4,696 detailed object instances. Then we combine these
descriptions to generate 211,320 subject-consistent image pairs. Through rigorous
quality control using GPT-4o, we selected 111,767 high-quality image pairs for
our final dataset. This extensive filtering process ensured the highest standards
of image quality and subject consistency, resulting in a collection of 223,534
high-quality training images.

    \section{Additional experimental results}

\subsection{Effect of training data}
\begin{figure}[t]
    \centering
    \includegraphics[width=\linewidth]{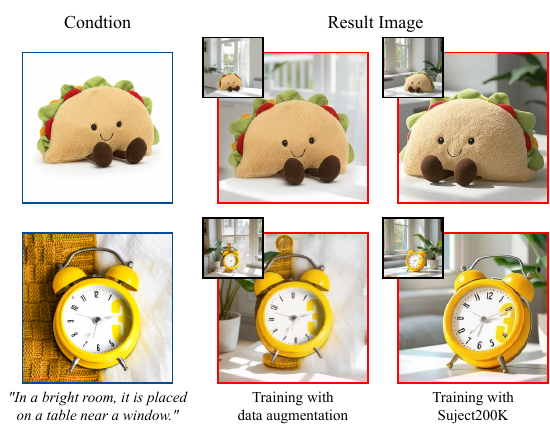}
    \caption{
    { Comparison of models trained with different data. The model trained by data augmentation tends to copy inputs directly, while model trained by our \datasetname generates novel views while preserving identity. }
    }
    \label{fig:data}
\end{figure}
For subject-driven generation, our model takes a reference image of a subject (e.g.,
a plush toy or an object) and a text description as input, aiming to generate novel
images of the same subject following the text guidance while preserving its key characteristics.

To validate the effectiveness of our \datasetname\ dataset described in Section~\ref{subsec:data},
we compare two training strategies for this task. The first approach relies on traditional
data augmentation, where we apply random cropping, rotation, scaling, and adjustments
to contrast, saturation, and color to the original images. The second approach
utilizes our \datasetname\ dataset. As shown in Figure~\ref{fig:data}, the model
trained with data augmentation only learns to replicate the input conditions
with minimal changes. In the first row, it simply places the taco plush toy in a
bright room setting while maintaining its exact appearance and pose. Similarly,
in the second row, the yellow alarm clock is reproduced with nearly identical details
despite the window-side placement instruction. In contrast, our \datasetname-trained
model demonstrates the ability to generate diverse yet consistent views of the subjects
while faithfully following the text prompts.

\subsection{Evaluation for subject-driven generation}
\label{subsec:ap:sub} \textbf{Framework and criteria.} To systematically
evaluate subject-driven generation quality, we establish a framework with five criteria
assessing both preservation of subject characteristics and accuracy of requested
modifications:
\begin{itemize}
    \item \textbf{Identity Preservation}: Evaluates preservation of essential identifying
        features (e.g., logos, brand marks, distinctive patterns)

    \item \textbf{Material Quality}: Assesses if material properties and surface
        characteristics are accurately represented

    \item \textbf{Color Fidelity}: Evaluates if colors remain consistent in regions
        not specified for modification

    \item \textbf{Natural Appearance}: Assesses if the generated image appears realistic
        and coherent

    \item \textbf{Modification Accuracy}: Verifies if the changes specified in the
        text prompt are properly executed
\end{itemize}

\begin{table*}
    [t]
    \centering
    \begin{tabular}{l|ccccc|c}
        \toprule Method                                                  & Identity      & Material      & Color         & Natural        & Modification  & Average       \\
                                                                         & preservation  & quality       & fidelity      & appearance     & accuracy      & score         \\
        \midrule \midrule \multicolumn{7}{c}{Average over 5 random seeds} \\
        \midrule IP-Adapter (SD 1.5)                                     & 29.4          & 86.1          & 45.3          & 97.9           & 17.0          & 55.1          \\
        SSR-Encoder                                                      & 46.0          & \textbf{92.0} & 54.2          & 96.3           & 28.5          & 63.4          \\
        IP-Adapter (FLUX)                                                & 11.8          & 65.8          & 30.8          & 98.1           & 57.7          & 52.8          \\
        \rowcolor{cvprblue!15} Ours                                      & \textbf{50.6} & 84.3          & \textbf{55.0} & \textbf{98.5}  & \textbf{75.8} & \textbf{72.8} \\
        \midrule \multicolumn{7}{c}{Best score over 5 random seeds}       \\
        \midrule IP-Adapter (SD 1.5)                                     & 56.3          & 98.9          & 70.1          & 99.7           & 37.2          & 72.5          \\
        SSR-Encoder                                                      & 64.3          & \textbf{99.2} & 74.4          & 99.1           & 53.6          & 78.1          \\
        IP-Adapter (FLUX)                                                & 27.5          & 86.1          & 53.6          & 99.9           & 74.9          & 68.4          \\
        \rowcolor{cvprblue!15} Ours                                      & \textbf{82.3} & 98.0          & \textbf{88.4} & \textbf{100.0} & \textbf{90.7} & \textbf{91.9} \\
        \bottomrule
    \end{tabular}
    \caption{Quantitative evaluation results (in percentage) across different
    evaluation criteria. Higher values indicate better performance.}
    \label{tab:quant_results}
\end{table*}

\begin{figure}[t]
    \centering
    \includegraphics[width=\linewidth]{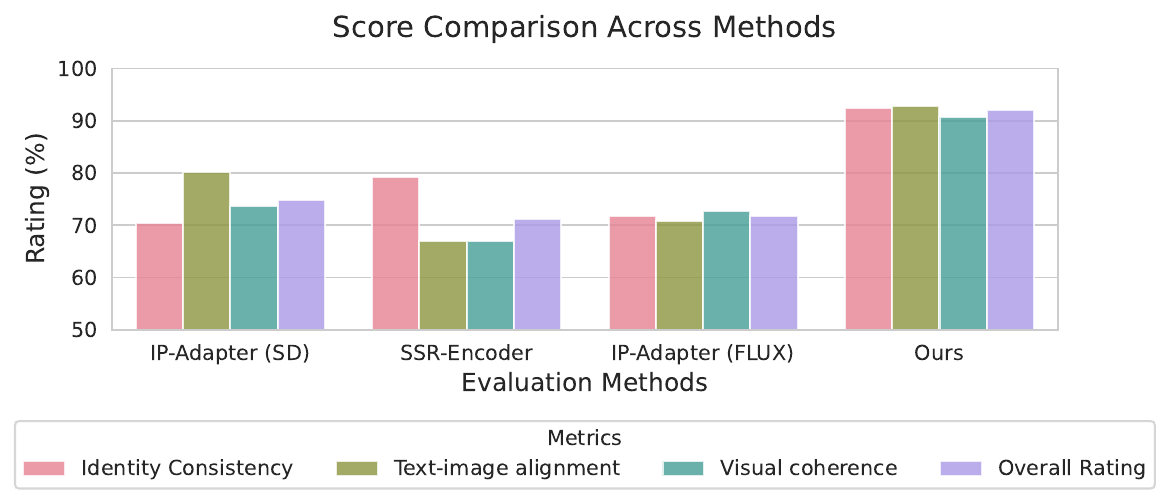}
    \caption{ User study results comparing different methods across three metrics:
    identity consistency, text-image alignment, and visual coherence. }
    \label{fig:user}
\end{figure}

\textbf{User studies.} To further validate our approach, we conducted user studies
collecting 375 valid responses. Participants evaluated the generated images
across three key dimensions: identity consistency, text-image alignment, and visual
coherence between subjects and backgrounds. The results shown in Figure~\ref{fig:user}
corroborate our quantitative findings, with our method achieving superior performance
across all evaluation criteria.

\subsection{Additional generation results}
\begin{figure*}[!t]
    \centering
    \includegraphics[width=\linewidth]{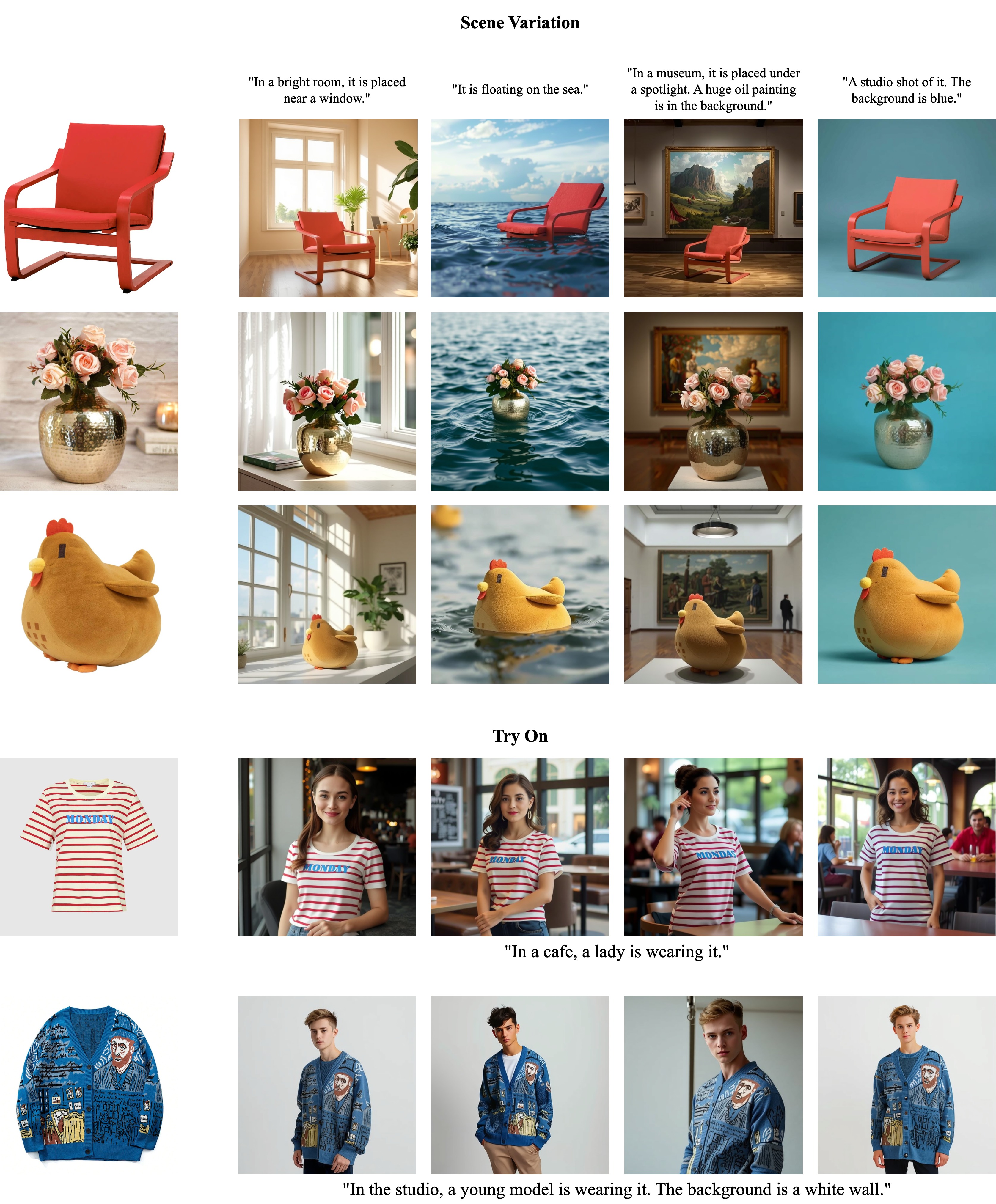}
    \caption{
    { More results on other subject-driven generation tasks. } }
    \label{fig:more}
\end{figure*}

We showcase more generation results from our method. Figure~\ref{fig:more_db}
presents additional results on the DreamBooth dataset, while Figure~\ref{fig:more}
demonstrates our method's effectiveness on other subject-driven generation tasks.

\begin{figure*}[!t]
    \centering
    \includegraphics[width=\linewidth]{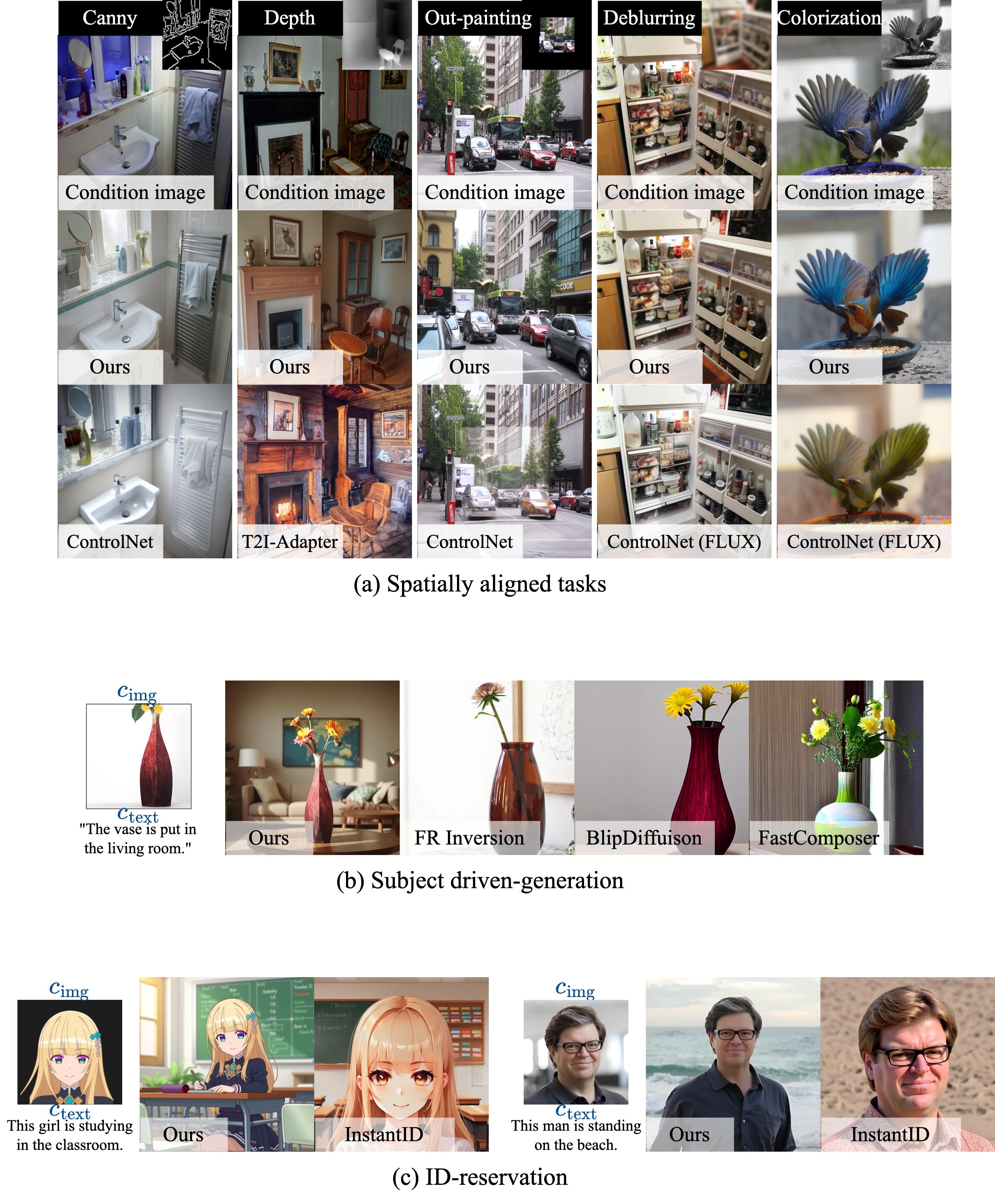}
    \caption{
    { More results on Dreambooth dataset. } }
    \label{fig:more_db}
\end{figure*}

\begin{figure*}[!t]
    \centering
    \includegraphics[width=\linewidth]{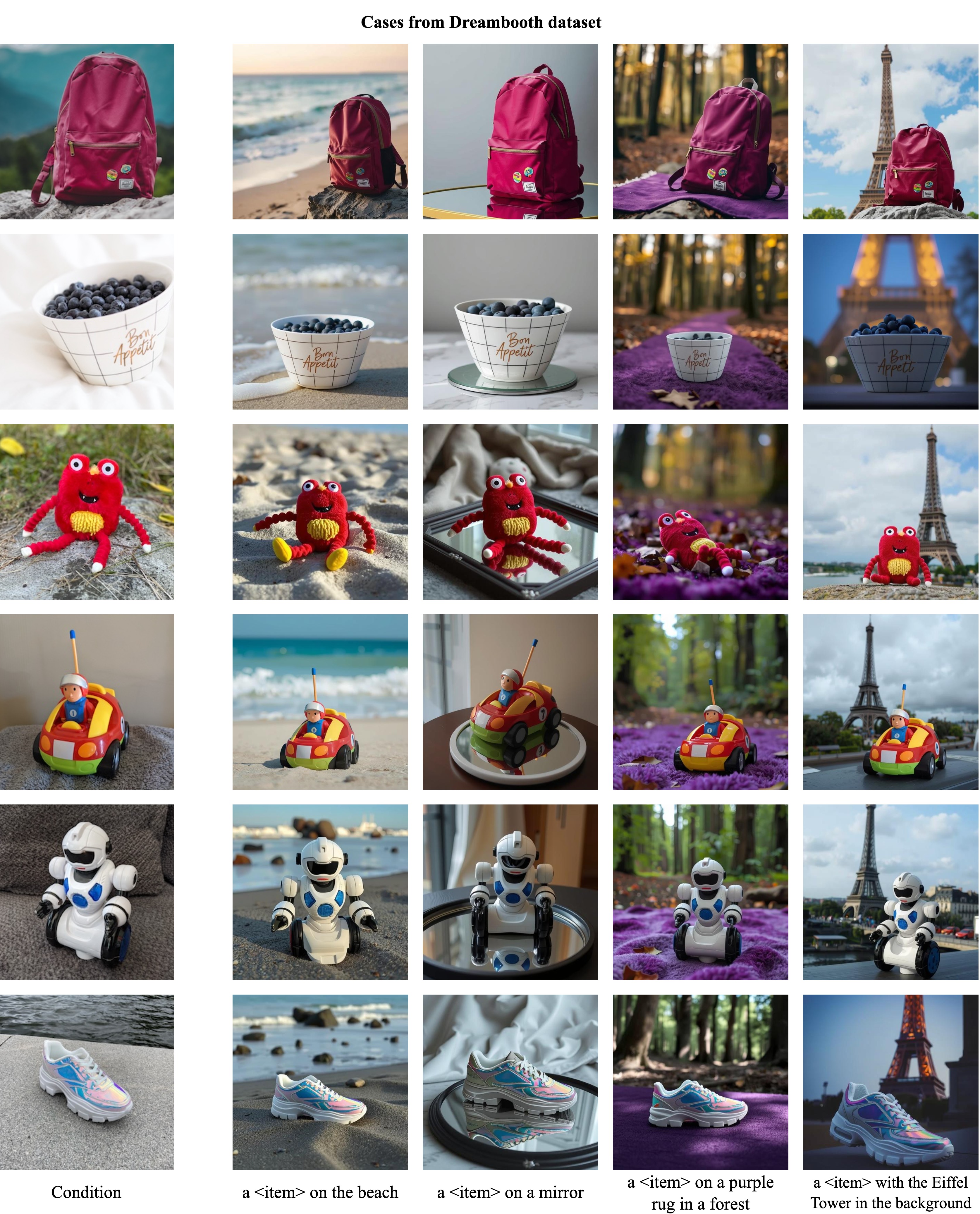}
    \caption{
    { More results on Dreambooth dataset. } }
    \label{fig:more_db}
\end{figure*}

\end{document}